\documentclass[conference]{IEEEtran}
\ifCLASSINFOpdf
\else
\fi
\usepackage{url}

\usepackage{xcolor}
\usepackage[pdftex]{graphicx}
\usepackage{amsfonts}
\usepackage{amsmath}
\usepackage{paralist}
\usepackage{cite}
\usepackage{wrapfig}
\usepackage[utf8]{inputenc}
\usepackage[english]{babel}
\usepackage{amsthm}
\usepackage{subfig}
\usepackage{graphicx}
\usepackage{enumitem}
\usepackage{mdframed}

\newtheorem{definition}{Definition}

\newcommand{\sectionname}{Section}

\begin{document}
%
\title{Process Mining Meets Causal Machine Learning: Discovering Causal Rules from Event Logs}


\author{
    \IEEEauthorblockN{Zahra Dasht Bozorgi\IEEEauthorrefmark{1}, 
    Irene Teinemaa\IEEEauthorrefmark{2},
    Marlon Dumas\IEEEauthorrefmark{3}, 
    Marcello La Rosa\IEEEauthorrefmark{1}, 
    Artem Polyvyanyy\IEEEauthorrefmark{1}}
    \IEEEauthorblockA{\IEEEauthorrefmark{1}University of Melbourne
    \\zdashtbozorg@student.unimelb.edu.au
    \\\{marcello.larosa, artem.polyvyanyy\}@unimelb.edu.au}
        \IEEEauthorblockA{\IEEEauthorrefmark{2}Booking.com
    \\irene.teinemaa@booking.com}
    \IEEEauthorblockA{\IEEEauthorrefmark{3}University of Tartu
    \\marlon.dumas@ut.ee}
}


%


\maketitle

\begin{abstract}
This paper proposes an approach to analyze an event log of a business process in order to generate case-level recommendations of treatments that maximize the probability of a given outcome. Users classify the attributes in the event log into controllable and non-controllable, where the former correspond to attributes that can be altered during an execution of the process (the possible treatments). We use an action rule mining technique to identify treatments that co-occur with the outcome under some conditions. Since action rules are generated based on correlation rather than causation, we then use a causal machine learning technique, specifically uplift trees, to discover subgroups of cases for which a treatment has a high causal effect on the outcome after adjusting for confounding variables. We test the relevance of this approach using an event log of a loan application process and compare our findings with recommendations manually produced by process mining experts.
\end{abstract}



%
\IEEEpeerreviewmaketitle

\section{Introduction}

\noindent
A business process is a collection of events, activities, and decisions that collectively lead to an outcome that can be of value to a customer~\cite{FBPM2}. Some outcomes are value-adding (e.g. the customer is satisfied with the delivery of a product) while others are not (e.g.\ a customer submits a complaint). Naturally, organizations strive to maximize the positive outcome rate of their processes or, conversely, to minimize the error rate.

Process mining techniques allow one to analyze the executions of a process to uncover sources of negative outcomes. Existing process mining techniques, such as \cite{DBLP:conf/bpm/LehtoHH16,DBLP:conf/simpda/LehtoHH17,DBLP:conf/dnis/GuptaAS15,DBLP:conf/bpm/SuriadiOAH12}, are geared towards identifying \textit{correlation} between observational data and outcomes (e.g.\ cases where the customer is satisfied have less rework loops) rather than \textit{causation} (e.g.\ customers who submit incorrect details cause more rework loops, leading to lower satisfaction). Meanwhile, causal inference techniques allow one to discover and measure causal relations between \emph{treatments} (e.g.\ checking the customer data) and \emph{outcomes} (e.g.\ the customer is satisfied) both from randomized experiments and from observational data. 

Recently, a family of techniques, namely causal machine learning, have emerged, which make use of machine learning methods to analyze causal effects. Causal machine learning encompasses techniques for estimating the causal effect of a treatment on an outcome given a set of potentially confounding variables (average treatment effect estimation) as well as techniques for classifying samples in a population based on the incremental effect of applying a treatment versus not applying it with respect to an outcome (uplift modeling).


In this study, we leverage these techniques to address the following question: Given a set of treatments (each with a certain cost), which may affect a business process outcome (with a certain benefit), which treatments yield the highest causal effect on the outcome and to which subset of cases should they be applied? In  line with this, the contribution of this paper is an approach to:
\begin{itemize}
    \item discover case-level treatment recommendations to increase the positive outcome rate of a process;
    \item identify subsets of cases to which a recommendation should be applied;    
    \item estimate the causal effect and the incremental Return-on-Investment (ROI) of a treatment.
\end{itemize}

The approach is designed to require minimal input from users. Users specify which attributes in the event log are controllable, meaning that their value can be manipulated by process participants, i.e. the employees who perform the various process tasks. Setting the value of a controllable attribute corresponds to a treatment. For example, in an order-to-cash process, setting an attribute \emph{discountGranted} to true means that a discount was granted. The attributes capturing such treatments may be derived during log pre-processing from the presence or absence of certain tasks, e.g.\ \emph{discountGranted} may be derived from the presence of task ``Grant Discount''. 

Given this input, we apply a technique to discover precondition-treatment-outcome rules with high support. Since neither a rule's support nor its confidence imply causation, we use a causal machine learning technique to assess the causal effect of the rule and to discover subsets of cases for which the treatment has the highest incremental success probability (\textit{uplift}). We then select the rules with the highest uplift. We report on a validation of this approach using a log of a process mining challenge (BPIC 2017) and compare the findings of our approach against those reported in the entries of this challenge.



The structure of the paper is as follows. We discuss related work in \sectionname~\ref{sec:relatedWork}. We introduce preliminary concepts in \sectionname~\ref{sec:preliminaries}, describe our approach in \sectionname~\ref{sec:approach} and present the validation in \sectionname~\ref{sec:evaluation}. In \sectionname~\ref{sec:conclusion} we conclude the paper and discuss future work.

\section{Related Work}
\label{sec:relatedWork}

\noindent
Previous work has shown that one can rely on influence analysis to identify improvement opportunities from event logs~\cite{DBLP:conf/bpm/LehtoHH16,DBLP:conf/simpda/LehtoHH17}. 
In \cite{DBLP:conf/dnis/GuptaAS15}, rules to describe root causes of anomalous process cases are extracted, while  \cite{DBLP:conf/bpm/SuriadiOAH12} relies on classical data mining methods to discover key attributes for root-cause analysis.
These methods identify correlations between attributes and the outcome but do not test for causality.

The problem of discovering cause-effect relations is addressed in \cite{DBLP:conf/caise/HompesMRDBA17}. This approach performs time series analysis to identify causal relations. This is different from our approach as the analysis is done at the process-level, while we provide case-level recommendations. The work in \cite{DBLP:conf/icsoc/BoseGCRD15} studies process-level factors that impact outcomes but fails to determine causalities between the two. 
A manual approach for confirming pre-identified causal relationships was proposed in \cite{DBLP:conf/bpm/NarendraA0D19}. 
It uses structural causal models to confirm cause-effect assumptions, control the effects of confounding, and answer counterfactual questions about the process.

Polyvyanyy~\emph{et~al}.~\cite{POLYVYANYY2019345} present a (semi-)automated approach, called \emph{causality mining}, to discover causal dependencies between events in large arrays of data.
The discovery is based on a notion of proximity of events in terms of time, space, and semantics.
The level of automation depends on the availability of the formalized domain knowledge.

In summary, previous work either addresses the problem of finding correlation rather than causation between factors and outcomes, or causation is addressed at the process-level, not at the case level. Moreover, previously identified causal effects either need to be confirmed manually, or extensive domain knowledge is required as input.

\section{Preliminaries}\label{sec:preliminaries}

\noindent
In this section we formalize preliminary concepts that are required to describe our approach, such as event logs, action rule mining, causal inference and uplift trees.
\subsection{Event Logs}
\noindent
Process mining studies methods for improving real-world processes based on event data \cite{DBLP:books/sp/Aalst16}. These data are often available in the form of an \emph{event log}. Event logs contain records of completed cases of a process. Each case is a record of the execution of a particular process instance and consists of a number of events. Each event has three mandatory attributes: (1) the case identifier indicating which case that event belongs to, (2) the activity name specifying the related activity for that event, and (3) the timestamp, showing when the event occurred. In addition, an event may have attributes, such as the resource carrying out the related activity. An event is formally defined as follows:
\begin{definition}(Event)
\emph{An \emph{event} is a tuple $(a,c,t, \langle(d_1,v_1),$ 
$\ldots,(d_m,v_m) \rangle)$, where $a$ is an activity name, $c$ is a case ID, $t$ is a timestamp, and $(d_1,v_1),\ldots,(d_m,v_m)$, $m\in \mathbb{N}$, are event attribute name-value pairs. Given an event $e$, $c_e$ denotes the identifier of the case.}
\end{definition}

A trace is a sequence of events that captures the execution of one case of a business process. An event log is defined as a set of traces.


\begin{definition}(Event Log)
\emph{ Let $\mathcal{E}$ be the universe of events. An \emph{event log} is a set $L \subset \mathcal{E}^*$.}
\end{definition}
\subsection{Action Rule Mining}
\noindent
Action rule mining is an extension of classification rule discovery \cite{DBLP:series/sci/2013-468}. While a classification rule predicts the class label of a data object, an action rule suggests what attribute values should be changed to increase the likelihood of that object being re-classified to another group. In \cite{DBLP:series/sci/2013-468}, action terms and action rules are defined as:

\begin{definition}(Atomic Action Terms)
\emph{An \emph{atomic action term} is an expression $(m: m_1 \to\ m_2)$, where $m$ is an attribute and $m_1$ and $m_2$ are possible values for attribute $m$.}
\end{definition}

If $m_1 = m_2$ then $m$ is \emph{uncontrollable}, denoted by $(m: m_1)$.

\begin{definition}(Action Terms)
\emph{A set of \emph{action terms} is the smallest set such that:
1. If $t$ is an atomic action term, then $t$ is an action term.
2. If $t_1$ and $t_2$ are action terms, then $t_1 \land\ t_2$ is an action term.
3. If an action term $t$ contains atomic action terms $(m: m_1 \to\ m_2)$ and $(n: n_1 \to\ n_2)$, then $m \neq n$.} 
\end{definition}

\begin{definition}(Action Rules)
\emph{An \emph{action rule} is an expression $r = [t_1 \Rightarrow\ t_2]$, where $t_1$ is an action term and $t_2$ is an atomic action term.}
\end{definition}


\subsection{Causal Inference}
\noindent
Causal inference is concerned with determining the causal effect between an intervention (treatment) and an outcome \cite{WhatIfBook}. Suppose that we have a treatment $A$ and an outcome $Y$. A potential outcome $Y^a$ is the outcome that \emph{would} be observed if the intervention was set to $A=a$. Focusing on a single intervention, each case in an event log has two potential outcomes, $Y^{a=1}$ for receiving the intervention and $Y^{a=0}$ for not receiving it. With these definitions, the average treatment effect (ATE) is defined as:



\begin{definition}(Average Treatment Effect)
\emph{Suppose we have an event log consisting of a number of cases. Let $\mathbb{E}[Y^{a=1}]$ be the average outcome if all cases in the log receive the treatment and $\mathbb{E}[Y^{a=0}]$ be the average outcome if all cases do not receive the treatment.
\begin{center}$\mathit{ATE}: \tau = \mathbb{E}[Y^{a=1}-Y^{a=0}]$\end{center} 
}
\end{definition}

Researchers are often interested in estimating the Conditional Average Treatment Effect (CATE), which is the expected treatment effect for a subgroup in the cases being studied. It enables personalizing treatments for each case and leads to a better understanding of causal mechanisms \cite{Kunzel4156}.

\begin{definition}(Conditional Average Treatment Effect) \emph{Suppose that X is a set of attributes characterizing the subgroup of interest. 
\begin{center}$\mathit{CATE}: \tau(x) = \mathbb{E}[Y^{a=1}-Y^{a=0}|X=x]$\end{center}
}
\end{definition}

Measuring the causal effect would be straightforward if we knew both potential outcomes for each case. However, in the real world, we can only observe one outcome for each case, corresponding to the treatment that the case actually received. 
To identify causal effects from observational data, three conditions must be met: \emph{Exchangeability}, \emph{Positivity}, and \emph{Consistency}.
Exchangeability (also known as ignorablility) means that given pre-treatment attributes $X$, treatment assignment is independent of the potential outcomes. 
\begin{center}$Y^1,Y^0 \perp \!\!\! \perp A|X$\end{center}
The consistency assumption states that the potential outcome under treatment $A=a$ is equal to the observed outcome if the actual treatment received is $A=a$.
\begin{center}$Y=Y^a$ if $A=a$ for all $a$\end{center}
Finally, the positivity assumption states that for every set of values for $X$, treatment assignment is not deterministic. This means that every subgroup of interest has some chance of getting either treatment.
\begin{center}$P(A=a|X=x)>0$ for all $a$ and for all $x$\end{center}

\begin{figure}[h]
\centering

   \includegraphics[scale=.5]{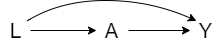}

\caption{Causal DAG depicting a confounding effect.}
\label{fig:confounding}
\vspace{-1mm}
\end{figure}

In practice, the most problematic of these three conditions is exchangeability. One approach to ensure that this condition is met is to conduct a randomized experiment (also known as an A/B test), where the treatment is assigned randomly to each case. Randomization of treatment assignment ensures that the treated and untreated groups are exchangeable, so that the causal effect can be consistently estimated from the observed data. However, conducting a randomized experiment is not always possible, since it might be expensive, time-consuming, or unethical. In these situations, the best we can do is to carry out an observational study.
In observational studies, treatment is often not randomized, meaning that the characteristics of the treated group might be different from the untreated. 
If the treatment assignment is not independent of the potential outcomes, then there exists a set of variables that affect both treatment and outcome. This is known as a confounder. Fig. \ref{fig:confounding} depicts the causal relations between a treatment $A$, an outcome $Y$, and a shared cause (i.e. a confounder) $L$. 

To estimate causal effects from observational data, we need to control confounding. To this end, we need to 
identify a set of variables, such that adjusting for these variables would make the exchangeability assumption hold. If some of the adjustment variables cannot be observed in the data, then causal effects are not identifiable in the observational study.

\subsection{Uplift Tree}
\noindent
Uplift modeling is concerned with estimating the causal effect of an action on the outcome of a particular instance (e.g. customer) \cite{DBLP:journals/kais/RzepakowskiJ12}. In other words, the aim is to estimate the change in class probabilities caused by an action. This is different from conventional prediction problems where a model is used to predict an outcome. For example, consider a marketing campaign. A conventional classifier would predict which customers will buy a product after a marketing action is performed without taking into account whether these users would have bought the product if the marketing action had not taken place. However, marketers are actually interested in finding the customers who are most likely to buy something \emph{because of} the marketing action. This is what uplift modeling is trying to achieve, which amounts to identifying subsets of instances with a high $\mathit{CATE}$.

In this study, we apply uplift modeling to business processes. We seek to estimate the change in the outcome of a process instance because of an action being applied to that case. Many uplift modeling approaches exist in the literature. We use the method proposed in \cite{DBLP:journals/kais/RzepakowskiJ12} to discover subgroups in the event log where a proposed treatment works best. It is a tree-based algorithm where the splitting criterion is designed to maximize the difference in $\mathit{CATE}$. 
The splitting criterion is the following:
\vspace{-1mm}
\begin{center}
    $D_{gain} = D_{\mathit{AfterSplit}}\left(P^T(Y) : P^C(Y) \right) - D_{\mathit{BeforeSplit}}\left( P^T(Y) : P^C(Y) \right)$,
\end{center}
\vspace{-1mm}
where $D$ can be substituted by the KL-divergence, squared Euclidean distance, or the chi-squared divergence and $P^T(Y)$ and $P^C(Y)$ are the probability distributions of the outcome in the treatment and control groups, respectively.

\section{Approach}\label{sec:approach}
\begin{figure*}[t]
  \includegraphics[width=\textwidth]{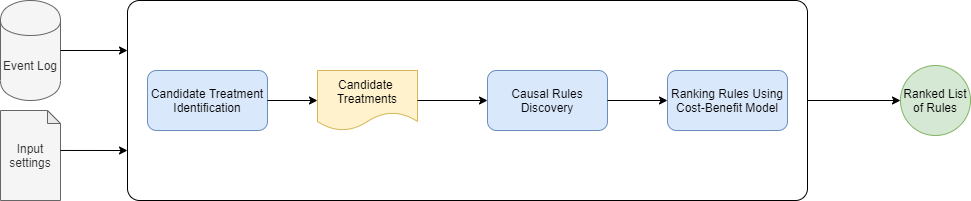}
  \caption{Overview of the approach.}
  \label{fig:approach}
\end{figure*}

\noindent
Our approach requires an event log as input along with input settings essential for the construction of action rules and uplift trees. It consists of three steps as shown in Fig. \ref{fig:approach}. First, candidate treatments are generated using action rule mining. Next, we identify subgroups in the population for every candidate treatment using uplift trees. Finally, we present a cost-benefit model to rank the rules based on the benefit of a positive outcome, the cost of treatment and its uplift.

\subsection{Candidate Treatments Identification}
\noindent
This step requires the user to provide an event log, the controllable and uncontrollable attributes, the outcome variable, and a minimum support. The classification of attributes into controllable and uncontrollable ensures that the candidate treatments are actionable, meaning that no change in the uncontrollable attributes is suggested by the action rules. In action rule mining, candidate treatment extraction is based on support. We seek to obtain rules that are likely to generate high revenue, which implies that the treatment should be related to the effect for a sufficiently large sub-population of cases. This is achieved by the support threshold. For example, the user may decide that a candidate treatment should be linked to the effect in at least 2\% of cases (support threshold); otherwise, the treatment is discarded.

\begin{figure*}[t]
    \begin{mdframed}[backgroundcolor=yellow!10]
        $\mathit{Rule}: r = [(\mathit{CreditScore}: \mathit{low}) \wedge (\mathit{NoOfTerms}: [6-48] \rightarrow [97-120])] \Longrightarrow \left[\mathit{Selected}: 0 \rightarrow 1 \right]$,\\ \emph{with support $0.057$ and confidence $0.764;$}
    \end{mdframed}
    \vspace{-1mm}
    \caption{Example Action Rule.}\label{actionRule}
\end{figure*}

Fig. \ref{actionRule} shows an example action rule. It was obtained based on the BPI Challenge 2017 event log using the method described in~\cite{DBLP:series/sci/2013-468}. 
The rule states that in cases where the customer's credit score is low, changing the number of terms from the interval 6--48 months to 97--120 months will increase the likelihood of the outcome variable (Selected) to change from negative (0) to positive (1). The rule is supplied with the support and confidence measurements. In the rule, \emph{CreditScore} is an example of an uncontrollable attribute, while the number of terms is considered controllable because the company can take steps to reduce or increase it. In our method, however, we only use the treatment part of the action rule. 
This is because action rules are an extension of classification rules, and they identify associations rather than causation. We seek to find a sub-population where the treatment causes the desired outcome. Therefore, in the next step, we use uplift modeling to discover \emph{causal} rules.

\subsection{Causal Rules Discovery}
\noindent
This step aims to discover subgroups $X$ for which a certain treatment $A$ has a high positive causal effect on the outcome $Y$. To this end, for each candidate treatment, we perform steps detailed below.

First, we build an uplift tree and take rules with high \emph{uplift}:
\begin{center}
$Pr(Y=1|A=1,X=x)-Pr(Y=1|A=0,X=x)$.
\end{center}

A popular but unjustified belief regarding uplift from observational data is that it can be estimated using the above formula. Uplift cannot be estimated this way unless we assume that for each case in the event log, $A$ is independent of the counterfactual outcomes $Y^1$ and $Y^0$ conditional on X (exchangeability assumption)~\cite{pmlr-v67-gutierrez17a}.
This assumption holds only when treatment assignment is randomized. In randomized controlled trials, treatment is randomized by design. However, in observational studies, the treated and the untreated cases are systematically different. The advantage of using the uplift tree method is that we can address the above issue by using its normalization feature. Normalization punishes tests that split the treatment and control groups in different proportions. These splits indicate situations where the test is not independent of the group assignment, and thus, the exchangeability assumption is violated. For a test $A$ and the KL-divergence criterion, normalizing factor is calculated as follows:

\begin{center}
    $I(A) = H\left(\frac{N^T}{N},\frac{N^C}{N}\right)\mathit{KL}\left(P^T(A):P^C(A)\right) + \frac{N^T}{N}H\left(P^T(A)\right) + \frac{N^C}{N}H\left(P^C(A)\right) + \frac{1}{2}$,
\end{center}
where $N^T$ and $N^C$ are the numbers of cases in the treatment group and the control group, respectively, and $N = N^T + N^C$ is the total number of cases.
For the squared Euclidean distance and the Chi-squared divergence, entropy is replaced by the Gini index. The first term of this factor punishes tests with imbalanced splits and, therefore, adjusts for confounding effects. The following two terms prevent bias towards tests with high numbers of outcomes. After normalization, the final splitting criterion is the gain divided by the normalizing value. 

While the normalization factor adjusts for biases related to observed confounders provided to the uplift tree as input variables, it does not ensure that there are no unobserved confounders that could invalidate the interpretation of the uplift estimates as true causal effects.
As an optional step, to strengthen the validity of the study, a manual check for confounding effects can be carried out by specifying a causal graph (depicting both observed and unobserved confounders) and identifying whether a valid adjustment set exists using the back-door criterion as described in~\cite{pearl2009causality}.

\begin{figure*}[t]
\vspace{-1mm}
\centering
  \includegraphics[width=\textwidth,height=6cm]{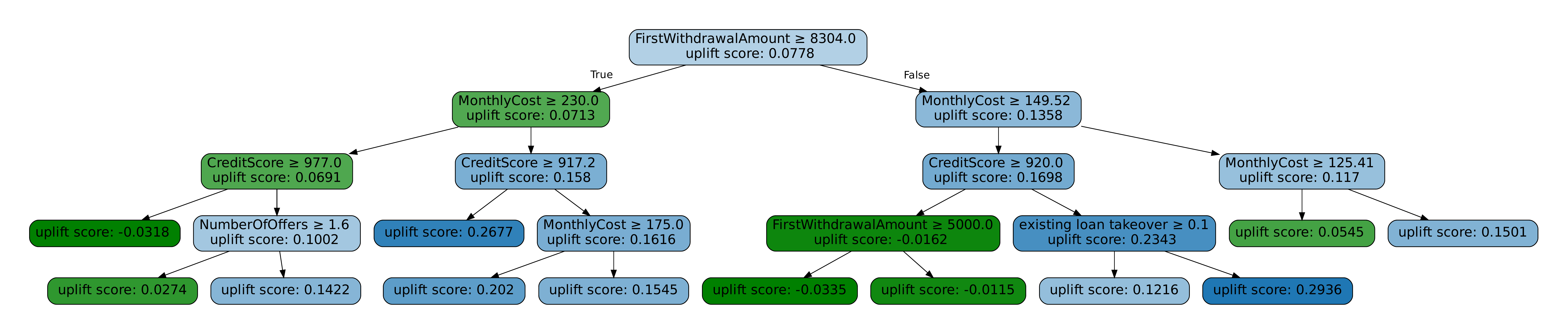}
  \caption{Example uplift tree.}\label{fig:uplift_tree}
\end{figure*}

Fig.~\ref{fig:uplift_tree} shows an example uplift tree for one of the rules extracted in the case study. The sub-populations of interest are in the leaves of the tree, and the user may pick the leaves that have an uplift score above a certain threshold.

\subsection{Ranking Rules Using a Cost-Benefit Model}
\noindent
An uplift score of an identified rule quantifies the causal effect of applying each intervention. However, it is not always profitable to carry out interventions with a negative incremental return-on-investment (ROI). So, we incorporate the uplift measure in a cost-benefit model that corresponds to the incremental ROI. Knowing the expected causal effect (estimated in the previous step) of applying a treatment to the discovered sub-population, the value of the desired outcome, and the costs of the treatments, we produce a cost-aware ranking of the rules. According to~\cite{DBLP:journals/corr/abs-1908-05372}, there are two types of treatment costs: 1) Fixed impression cost, i.e. the cost that occurs when applying the treatment, such as the cost of a phone call, 2) Triggered cost, i.e. the cost that occurs only if the treated case reaches a positive outcome such as lowering the interest rate of a loan. In the following, we assume that the triggered costs are not present.

We use this notation to define the cost-benefit model:
\begin{itemize}
    \item $v$: value (benefit) of a positive outcome, assuming that it is constant and given as prior knowledge; 
    \item $c$: impression cost for a treatment; 
    \item $u$: uplift of applying a treatment; and
    \item $n$: size of the treated group.
\end{itemize}

We define the net value of applying a treatment is as follows:

\begin{center}
    $net = n \times (u \times v - c)$.
\end{center}

Note that $n \times u \times v$ represents the incremental value of the treatment and $n \times c$ the incremental cost.

\section{Evaluation}
\label{sec:evaluation}
\noindent
The proposed approach was implemented in Python 3.7 using the ActionRules\footnote{\url{https://github.com/lukassykora/actionrules}} package for generating actionable recommendations and the CausalML package~\cite{chen2020causalml} for constructing uplift trees. The relevance of the approach is shown through a case study using the BPI Challenge (BPIC) 2017 log.\footnote{doi:10.4121/uuid:5f3067df-f10b-45da-b98b-86ae4c7a310b
} We chose this log among all other BPIC logs because the approach requires a setting where an outcome can be influenced by interventions that can take the form of a change in the case attributes. The BPIC 2017 log was the only one that satisfies these criteria. Since we did not have access to a subject matter expert from the company that provided the log, we compare our results with the reports of the winners of the challenge. The main goal of this experiment is thus to compare the recommendations that we generated automatically with those that the winners of the challenge produced. 
\subsection{Dataset}
\noindent
The BPIC 2017 log records cases of a loan application process at a Dutch financial institute filed in 2016 and handled up until 2 February 2017. It contains 31,509 applications (cases), 1,202,267 events and 42,995 offers. In addition to the attributes found in the log, we engineered the number of offers made to the customer as an extra feature. We also filtered the log to remove cases where the value of the outcome variable was missing.

For each application, one or more offers may be created, but the customer may only select one offer. In many cases, the customer does not select any of the bank offers. Hence, the target variable in this study is the attribute `Selected'. It is a Boolean attribute that is equal to true if the customer selects an offer and false otherwise.

Next, we classified the other attributes into controllable and uncontrollable. These features were considered uncontrollable: 
\begin{itemize}
    \item Application type (new credit or limit raising);
    \item Loan goal (reason for the loan application);
    \item Customer credit score; and
    \item Requested amount.
\end{itemize}

The following features were classified as controllable: \smallskip 

\begin{itemize}
  \item Number of offers;
  \item Number of payback terms (months);
  \item Monthly cost; and
  \item Initial withdrawal amount.
\end{itemize}
\subsection{Results}
\noindent
We ran the action rule discovery algorithm on the above dataset with support = 3 and confidence = 55, resulting in 24 rules containing 17 distinct recommendations. For each recommendation, we constructed an uplift tree to find the sub-population for each actionable recommendation using the following settings:

\begin{itemize}
  \item maximum depth of the tree = 5;
  \item minimum number of samples for a split  = 200;
  \item minimum number of samples in the treatment group for a split  = 50;
  \item regularization parameter = 100;
  \item evaluation function = Kullback-Leibler divergence.
\end{itemize}

We extracted the following rules:

\begin{itemize}
  \item Action 1: Decreasing the initial withdrawal amount from 7,500--9,895 to 0--7,499.
  Sub-population: Limit raising customers with a credit score greater than 885 and a monthly cost below 120 Euros.
  \item Action 2: Increasing number of terms from 6--48 months to more than 120 months.
  Sub-population: Customers whose credit scores are between 899 and 943 and their first withdrawal amount is less than 8,304.
  \item Action 3: Increasing number of terms from 6--48 months to 61--96 months. 
  Sub-population: Customers whose loan goal is not existing loan takeover, have a credit score less than 920, their offer includes a monthly cost greater than 149 and the first withdrawal amount is less than 8,304. 
  \item Action 4: Increasing number of terms from 6--48 months to 97--120 months. 
  Sub-population 1: Customers with a credit score less than 982, first withdrawal amount greater than 8,304, and a monthly cost between 154 and 205.
  Sub-population 2: Customers with a credit score between 781 and 982, first withdrawal amount less than 8,304, and a monthly cost greater than 147.
  \item Action 5: Decreasing first withdrawal amount from 7,500--9,895 to 5,750--7,499 and decreasing number of terms if greater than 120 months to 49--60 months.
  Sub-population: Customers with an offer that has a monthly cost less than 150.
  \item Action 6: Decreasing first withdrawal amount from 7,500--9,895 to 0--7,499 and decreasing number of terms if greater than 120 months to 97--120 months.
  Sub-population: New credit applicants with a credit score less than 914 that have an offer with a monthly cost more than 150.
  \item Action 7: Increasing first withdrawal amount from 7,500--9,895 Euros to 9,896--75,000.
  Sub-population: New credit application where the loan goal is existing loan takeover and the customer credit score is 825. 
  \item Action 8: Decreasing first withdrawal amount from 9,896--75,000 to 1,490--7,499 and increasing number of terms from 49--60 to 97--120.
  Sub-population: Customers with credit score lower than 933 and monthly cost greater than 154.
\end{itemize}

\subsection{Discussion}
\noindent 
The BPI challenge had three categories: student, professional and academic. Below, we discuss the recommendations of the winners in each category and compare them with our findings. Since our recommendations are at case level, we only discuss the case-level recommendations in these reports.

\smallskip
\subsubsection{Academic}
\noindent
The winning report in this category~\cite{rodrigues2017stairway} starts by producing the as-is process model of the underlying loan application process. It then identifies the variants of the process in order to understand the data at hand better. The authors analyze the process outcome and recommend decreasing the monthly cost or increasing the number of terms to more than 120. This is very similar to our Action~2 in the previous section. However, we found that this rule should be applied to customers in a specific range of credit scores (between 899 and 943) and the first withdrawal amount (less than 8,304). It is for this specific sub-population that this action has a high incremental effect. For example, in the rule's uplift tree, it is indicated that if the first withdrawal amount is higher than 10,000, the uplift is only 8\%. Thus, while it might generally be beneficial to increase the number of terms, applying this action might not lead to an increase in revenue in all the cases.

The authors of~\cite{rodrigues2017stairway} also carried out an analysis on the application type. There are two application types in this dataset: new credit and limit raise. The majority of the applications are new credit applications (89\%). They found that applications for limit raising have a higher rate of success than new credit applications. Limit raising applications are included in a sub-population in only one rule. One possible reason for this is that this type of application already has a high likelihood of success. Thus, applying a treatment would be unnecessary. In addition, we have found that more than the application type, it is the credit score that determines the causal effect of an action on the case outcome. According to our results, credit score is included in almost all the rules, but the application type is present in less than half the rules.

Regarding the number of offers, the authors of the report observed that there is an association between having more offers and the customer not cancelling the application. The candidate treatment identification part of our method was not able to recommend any action regarding the number of offers made to the customer. This is because the action rules method generates rules based on support. In this dataset, one offer was created for the majority of applications. Therefore, any rule with a higher number of offers would not reach the minimum support threshold.

\smallskip
\subsubsection{Professional}
\noindent
Similar to the academic category, in this report~\cite{blevi2017process}, the authors discovered the association between the number of offers and a successful outcome and made the recommendation to increase the number of offers. They also analyzed different loan goals and concluded that it would be beneficial to improve the instructions on the required documentation. This is to decrease the duration of the application. They were able to identify the cause of the delays in the application process by manually checking the time needed to validate the application and the number of times requests for document completions are sent.

Further, they analyzed the impact of credit score on the customer's decision and found that high credit customers have a significantly higher chance of a successful outcome. This is in line with our finding: Actions 2--4 and 6--8 are concerned with customers with low credit scores, meaning that there is no need for change in applications with high credit scores.

The authors performed a predictive analysis to determine which attributes have an impact on the outcome. They found that ``credit score has a significant impact on the decision of the customer to take the offer or not''. They state that this might be because the bank has more competitive offers for higher credit customers, and they recommend the bank to investigate this issue further. This is in line with our finding that most of the discovered sub-groups in our rules include lower credit score customers who are more in need of treatments. 
However, we go beyond the finding in the report and provide concrete, actionable rules for customers with lower credit scores and are more likely to reject their offers.

Finally, the authors of the winning report in the Professional category found that all other variables being equal, raise limit applications are more likely to be successful than new credit applications. However, this finding is based on application type having a low p\_value, which indicates association rather than causation. Furthermore, they found that monthly cost and duration also impact the outcome. Regarding the monthly cost, this is related to the number of terms. So, if a rule recommends that the number of terms should be increased, it is implicitly recommending the monthly cost to be decreased. The duration of an application depends both on what the client nominated in their loan application and what the bank has approved. It also depends on the time it takes for the client to respond to the bank's requests (e.g.\ requests for additional documents) and the time for the bank to process the application. For the above reasons, these are spurious results, and, as such, they did not emerge in our findings.

\smallskip
\subsubsection{Student}
\noindent
In this report~\cite{povalyaeva2017bpic}, the authors provide five recommendations:
\begin{itemize}
  \item Send offers to clients as soon as possible.
  \item Maximize the number of terms.
  \item Minimize the first withdrawal amount.
  \item Minimize the monthly cost.
  \item Minimize the time intervals between sending the offers to the client.
\end{itemize}

The authors found that if offers are sent within four days of submitting the application, the cancellation rate may decrease between 5\% and 10\%. Moreover, they found that in the cases where multiple offers where made, keeping the mean time between the offers to below four days may decrease the cancellation rate by 2\% to 5\%. Our recommendations do not include the timing of the offers because it requires additional feature engineering. Since we did very minimal feature engineering to keep the method as automated as possible, we cannot comment on the authors' first and last recommendations.

Regarding the number of terms, they found that by keeping it to above 60, they can decrease the cancellation rate by 3\% to 9\%. We also identified three rules regarding the number of terms (Actions 2--4 and 8). This recommendation is generally made for all the customers of the bank. However, we found that the magnitude of the increase has different causal effects depending on the circumstance of each case. For instance, suppose we have a customer with a credit score of 900. According to our Action 2, if the first withdrawal amount for this customer was determined to be less than 8,304, then the number of terms for this customer should be increased to more than 120. However, if the same customer has an offer with the first withdrawal amount greater than 8,304 and a low monthly cost, then the increase in the number of terms to the interval 97--120 is sufficient. Additionally, an overall increase in the number of terms will not necessarily be beneficial. For example, according to the uplift tree for our Action 3 (see Fig. \ref{fig:uplift_tree}), increasing the number of terms for clients with their loan goal being that of existing loan takeover, has a low uplift score. This means that for these types of applicants, offers with a high number of terms will not influence the outcome. It should also be noted that we have discovered two rules (Actions 5 and 6) where decreasing the number of terms is recommended, but only when it is paired with decreasing the first withdrawal amount. Rule 5 describes situations where the offer has a high initial withdrawal amount and a low monthly cost. So it recommends a more balanced offer where the customer can pay less initially but is charged slightly more monthly. Rule 6 describes a similar situation, but for bigger loans (e.g.\ when the monthly cost is higher than 150). 

According to the same report, the first withdrawal amount should be less than 5,000 Euros. By keeping it below this amount, a decrease of cancellation by 9\% to 12\% can be reached. We also found that in most of our rules, the initial withdrawal amount should be decreased. Only in one rule (Action 7), we found that it should be increased, namely for cases where the application type is new credit, loan goal is existing loan takeover, and the customer's credit score is 825. This might be due to the fact that in such applications, the customer does not have a high enough credit to demand a better offer and at the same time they may not want to increase the number of payback months, because they have already been paying for a previous loan. So, a high initial withdrawal amount might be attractive to such clients.

The authors' recommendation regarding the monthly cost is to minimize it. They found that having a monthly cost below 400 Euros will bring the cancellation rate down by 3\%. The monthly cost is closely related to the loan amount and the number of terms. With the loan amount being fixed, increasing the number of terms will automatically result in the monthly cost being decreased. So in those rules where we recommend to increase the number of terms (Actions 2--4 and 8), we are implicitly recommending the monthly cost to be decreased.

\subsection{Threats to validity.} 
\noindent
The above validation comes with the usual threat to external validity associated with case studies (lack of generalizability). We do not claim that the approach can discover relevant rules in other contexts. The validation also comes with threats to internal validity. The results may be affected by data quality issues in the event logs as well as misinterpretations of the semantics of the data or of the usefulness of the discovered rules. The latter threat is mitigated by the fact that we cross-checked our interpretation of the data and the results against the reports of the winners of the BPI Challenge who, at the time the contest took place, were able to indirectly resolve doubts about the data and its semantics with domain experts. To compensate for these threats, we provide a software artifact to enable other researchers to reproduce our experiments and to replicate them in other case studies (see link at the end of the paper). The fact that the experiments are based on observational data creates a potential threat to construct validity. The estimated uplift scores may not match those observed when the rules are deployed in practice. A rigorous A/B test should be conducted prior to deploying the recommendation rules in an operational setting.

\section{Conclusion and future work}
\label{sec:conclusion}
\noindent
We proposed an approach for analyzing event logs in order to generate recommendations for applying treatments during the execution of a case so as to maximize the probability of an outcome. The approach leverages causal machine learning techniques to estimate the causal effect of a treatment and to identify subsets of cases for which a treatment has the highest incremental effect (uplift). We sketched how the incremental ROI of applying a treatment may be derived from the uplift.

We validated the approach by applying it to the event log of the BPI Challenge 2017. Since we did not have access to a domain expert from the company that supplied the log, we compared our findings to those of the winners of the challenge. We found that most of the generated recommendations matched those of the winners. Furthermore, for each recommendation, our approach identified specific subsets of cases (e.g.\ specific types of loan applications or customers) for which the treatment could be most effective. 

The approach adjusts the estimation of the causal effect of a treatment against the variables extracted from the event log (e.g. case and event attributes). However, it does not adjust for exogenous confounding effects not explicitly captured in the log. An avenue for future work is to extend the approach with a method to identify contextual variables, such as time of the day, geographic location of the process stakeholders, weather, and to validate the discovered causal relations with respect to such variables, for example using structural models as in~\cite{DBLP:conf/bpm/NarendraA0D19}.

The reported validation is preliminary and does not involve feedback from potential users of the technique (managers, analysts) or external validation, for example, via A/B testing of the identified treatments. Conducting complementary validations of the proposed approach is another direction for future work.

\section*{Reproducibility}
\noindent
The source code and documentation to reproduce the experiments are available at \url{https://github.com/zahradbozorgi/CausalRulesDiscovery}
\section*{Acknowledgments}
\noindent 
Research funded by the Australian Research Council (grant DP180102839) and the European Research Council (PIX Project). Thanks to Tomáš Kliegr and Lukáš Sýkora for providing their Action Rules Jupyter notebook and to Mahmoud K. Shoush for his help with data preprocessing.



\bibliographystyle{IEEEtran}
%
\bibliography{lit}

\begin{thebibliography}{10}
\providecommand{\url}[1]{#1}
\csname url@samestyle\endcsname
\providecommand{\newblock}{\relax}
\providecommand{\bibinfo}[2]{#2}
\providecommand{\BIBentrySTDinterwordspacing}{\spaceskip=0pt\relax}
\providecommand{\BIBentryALTinterwordstretchfactor}{4}
\providecommand{\BIBentryALTinterwordspacing}{\spaceskip=\fontdimen2\font plus
\BIBentryALTinterwordstretchfactor\fontdimen3\font minus
  \fontdimen4\font\relax}
\providecommand{\BIBforeignlanguage}[2]{{%
\expandafter\ifx\csname l@#1\endcsname\relax
\typeout{** WARNING: IEEEtran.bst: No hyphenation pattern has been}%
\typeout{** loaded for the language `#1'. Using the pattern for}%
\typeout{** the default language instead.}%
\else
\language=\csname l@#1\endcsname
\fi
#2}}
\providecommand{\BIBdecl}{\relax}
\BIBdecl

\bibitem{FBPM2}
M.~Dumas, M.~L. Rosa, J.~Mendling, and H.~A. Reijers, \emph{Fundamentals of
  Business Process Management, 2nd Ed.}\hskip 1em plus 0.5em minus 0.4em\relax
  Springer, 2018.

\bibitem{DBLP:conf/bpm/LehtoHH16}
T.~Lehto, M.~Hinkka, and J.~Hollm{\'{e}}n, ``Focusing business improvements
  using process mining based influence analysis,'' in \emph{Proc. of BPM}, ser.
  LNBIP, vol. 260.\hskip 1em plus 0.5em minus 0.4em\relax Springer, 2016.

\bibitem{DBLP:conf/simpda/LehtoHH17}
T.~Lehto, M.~Hinkka, and J.~Hollmen, ``Focusing business process lead time
  improvements using influence analysis,'' in \emph{Proc. of SIMPDA)}, ser.
  {CEUR} Workshop Proceedings, vol. 2016.\hskip 1em plus 0.5em minus
  0.4em\relax CEUR-WS.org, 2017.

\bibitem{DBLP:conf/dnis/GuptaAS15}
N.~Gupta, K.~Anand, and A.~Sureka, ``Pariket: Mining business process logs for
  root cause analysis of anomalous incidents,'' in \emph{Proc. of DNIS
  Workshop}, ser. LNCS, vol. 8999.\hskip 1em plus 0.5em minus 0.4em\relax
  Springer, 2015.

\bibitem{DBLP:conf/bpm/SuriadiOAH12}
S.~Suriadi, C.~Ouyang, W.~M.~P. van~der Aalst, and A.~H.~M. ter Hofstede,
  ``Root cause analysis with enriched process logs,'' in \emph{Proc. of BPM
  Workshops}, ser. LNBIP, vol. 132.\hskip 1em plus 0.5em minus 0.4em\relax
  Springer, 2012.

\bibitem{DBLP:conf/caise/HompesMRDBA17}
B.~F.~A. Hompes, A.~Maaradji, M.~L. Rosa, M.~Dumas, J.~C. A.~M. Buijs, and
  W.~M.~P. van~der Aalst, ``Discovering causal factors explaining business
  process performance variation,'' in \emph{Proc. of CAiSE}, ser. LNCS, vol.
  10253.\hskip 1em plus 0.5em minus 0.4em\relax Springer, 2017.

\bibitem{DBLP:conf/icsoc/BoseGCRD15}
R.~P. J.~C. Bose, A.~Gupta, D.~Chander, A.~Ramanath, and K.~Dasgupta,
  ``Opportunities for process improvement: {A} cross-clientele analysis of
  event data using process mining,'' in \emph{Proc. of ICSOC}, ser. LNCS, vol.
  9435.\hskip 1em plus 0.5em minus 0.4em\relax Springer, 2015.

\bibitem{DBLP:conf/bpm/NarendraA0D19}
T.~Narendra, P.~Agarwal, M.~Gupta, and S.~Dechu, ``Counterfactual reasoning for
  process optimization using structural causal models,'' in \emph{Proc. of BPM
  Forum}, ser. LNBIP, vol. 360.\hskip 1em plus 0.5em minus 0.4em\relax
  Springer, 2019.

\bibitem{POLYVYANYY2019345}
A.~Polyvyanyy, A.~Pika, M.~T. Wynn, and A.~H. {ter Hofstede}, ``A systematic
  approach for discovering causal dependencies between observations and
  incidents in the health and safety domain,'' \emph{Safety Science}, vol. 118,
  2019.

\bibitem{DBLP:books/sp/Aalst16}
W.~M.~P. van~der Aalst, \emph{Process Mining - Data Science in Action, Second
  Edition}.\hskip 1em plus 0.5em minus 0.4em\relax Springer, 2016.

\bibitem{DBLP:series/sci/2013-468}
A.~Dardzinska, \emph{Action Rules Mining}, ser. Studies in Computational
  Intelligence.\hskip 1em plus 0.5em minus 0.4em\relax Springer, 2013, vol.
  468.

\bibitem{WhatIfBook}
M.~A. Hernan and J.~M. Robins, \emph{Causal Inference: What If}.\hskip 1em plus
  0.5em minus 0.4em\relax Boca Raton: Chapman \& Hall CRC, 2020.

\bibitem{Kunzel4156}
S.~R. K{\"u}nzel, J.~S. Sekhon, P.~J. Bickel, and B.~Yu, ``Metalearners for
  estimating heterogeneous treatment effects using machine learning,''
  \emph{Proc. of the National Academy of Sciences}, vol. 116, no.~10, 2019.

\bibitem{DBLP:journals/kais/RzepakowskiJ12}
P.~Rzepakowski and S.~Jaroszewicz, ``Decision trees for uplift modeling with
  single and multiple treatments,'' \emph{Knowl. Inf. Syst.}, vol.~32, no.~2,
  2012.

\bibitem{pmlr-v67-gutierrez17a}
P.~Gutierrez and J.-Y. Gérardy, ``Causal inference and uplift modelling: A
  review of the literature,'' in \emph{Proc. of Int. Conf. on Predictive
  Applications and APIs}, ser. Proc. of Machine Learning Research,
  vol.~67.\hskip 1em plus 0.5em minus 0.4em\relax PMLR, 2017.

\bibitem{pearl2009causality}
J.~Pearl, \emph{Causality}.\hskip 1em plus 0.5em minus 0.4em\relax Cambridge
  university press, 2009.

\bibitem{DBLP:journals/corr/abs-1908-05372}
Z.~Zhao and T.~Harinen, ``Uplift modeling for multiple treatments with cost
  optimization,'' \emph{CoRR}, vol. abs/1908.05372, 2019.

\bibitem{chen2020causalml}
H.~Chen, T.~Harinen, J.-Y. Lee, M.~Yung, and Z.~Zhao, ``Causalml: Python
  package for causal machine learning,'' 2020.

\bibitem{rodrigues2017stairway}
A.~Rodrigues, C.~Almeida, D.~Saraiva, F.~Moreira, G.~Spyrides, G.~Varela,
  G.~Krieger, I.~Peres, L.~Dantas, M.~Lana \emph{et~al.}, ``Stairway to value:
  mining a loan application process,'' \emph{BPI Challenge}, 2017.

\bibitem{blevi2017process}
L.~Blevi, L.~Delporte, and J.~Robbrecht, ``Process mining on the loan
  application process of a dutch financial institute,'' \emph{BPI Challenge},
  2017.

\bibitem{povalyaeva2017bpic}
E.~Povalyaeva, I.~Khamitov, and A.~Fomenko, ``Bpic 2017: density analysis of
  the interaction with clients,'' \emph{BPI Challenge}, 2017.

\end{thebibliography}

\end{document}